\def\paperID{46}
\def\confName{CVPR}
\def\confYear{2024}

\def\authorBlock{
    Rafael Elberg \qquad
    Denis Parra \qquad
    Mircea Petrache \\
    PUC Chile \\
    {\tt\small \{rafael.elberg, denis.parra, mpetrache\}@uc.cl}
}

\newif\ifreview 
\newif\ifarxiv \newcommand{\arxiv}{\arxivtrue}
\newif\ifcamera 
\newif\ifrebuttal 

\arxiv
\pdfoutput=1
\documentclass[10pt,twocolumn,letterpaper]{article}
\ifreview \usepackage[review]{cvpr} \fi
\ifarxiv \usepackage[pagenumbers]{cvpr} \fi
\ifrebuttal \usepackage[rebuttal]{cvpr} \fi
\ifcamera \usepackage{cvpr} \fi

\usepackage{graphicx}	
\usepackage{amsmath}	
\usepackage{amssymb}	
\usepackage{booktabs}
\usepackage{times}
\usepackage{microtype}
\usepackage{epsfig}
\usepackage[table,xcdraw,dvipsnames]{xcolor}
\usepackage{caption}
\usepackage{float}
\usepackage{placeins}
\usepackage{color, colortbl}
\usepackage{stfloats}
\usepackage{enumitem}
\usepackage{tabularx}
\usepackage{xstring}
\usepackage{multirow}
\usepackage{xspace}
\usepackage{url}
\usepackage{subcaption}
\usepackage{xcolor}
\usepackage[hang,flushmargin]{footmisc}

\ifcamera \usepackage[accsupp]{axessibility} \fi

% ...

\ifarxiv  \fi

\newcommand{\R}[1]{{%
    \textbf{%
        \ifstrequal{#1}{1}{\textcolor{red}{R#1}}{%
        \ifstrequal{#1}{2}{\textcolor{blue}{R#1}}{%
        \ifstrequal{#1}{3}{\textcolor{magenta}{R#1}}{%
        \ifstrequal{#1}{4}{\textcolor{teal}{R#1}}{%
                           \textcolor{cyan}{R#1}%
        }}}}%
    }%
}}

\DeclareMathOperator{\EX}{\mathbb{E}}

\usepackage{xr-hyper}

\makeatletter
\newcommand*{\addFileDependency}[1]{
  \typeout{(#1)}
  \@addtofilelist{#1}
  \IfFileExists{#1}{}{\typeout{No file #1.}}
}

\makeatother
\newcommand*{\myexternaldocument}[1]{
    \externaldocument{#1}
    \addFileDependency{#1.tex}
    \addFileDependency{#1.aux}
}

\definecolor{cvprblue}{rgb}{0.21,0.49,0.74}
\usepackage[pagebackref,breaklinks,colorlinks,citecolor=cvprblue]{hyperref}
\usepackage[capitalize]{cleveref}
\crefname{section}{Sec.}{Secs.}
\crefname{table}{Table}{Tables}
\crefname{figure}{Fig.}{Figs.}

\ifarxiv \crefname{appendix}{App.}{Apps.}
\else \crefname{appendix}{Suppl.}{Suppls.} \fi

\unless\ifarxiv \myexternaldocument{_supplementary} \fi

\begin{document}
\captionsetup[subfigure]{labelformat=empty}
\title{Long Tail Image Generation Through Feature Space Augmentation and Iterated Learning}
\author{\authorBlock}
\maketitle

\begin{abstract}
Image and multimodal machine learning tasks are very challenging to solve in the case of poorly distributed data. In particular, data availability and privacy restrictions exacerbate these hurdles in the medical domain. 
The state of the art in image generation quality is held by Latent Diffusion models, making them prime candidates for tackling this problem. However, a few key issues still need to be solved, such as the difficulty in generating data from under-represented classes and a slow inference process. To mitigate these issues, we propose a new method for image augmentation in long-tailed data based on leveraging the rich latent space of pre-trained Stable Diffusion Models. We create a modified separable latent space to mix head and tail class examples. We build this space via Iterated Learning of underlying sparsified embeddings, which we apply to task-specific saliency maps via a K-NN approach. Code is available at \url{https://github.com/SugarFreeManatee/Feature-Space-Augmentation-and-Iterated-Learning}
\end{abstract}
\section{Introduction}
\label{sec:intro}
With the rise of large multimodal models \cite{Fei2022, wu2023nextgpt, li2023multimodal} and Latent Diffusion models \cite{rombach2022highresolution, 10.1007/978-3-031-18576-2_12}, both image analysis and generation tasks have become dependent on the availability and quality of balanced training data \cite{parrots}. Due to the representation capabilities of these large models with several million, or even billions of parameters, volumes of data in similar magnitudes are required to avoid issues such as overfitting or learning undesirable biases. 

Obtaining extensive and well-distributed data is not an option for several essential domains. For example, for the analysis and generation of medical images, obtaining data can be complex (as they are subject to patient confidentiality) and expensive (as they are tied to real-world medical procedures and imaging exams) \cite{GARCEA2023106391}. Moreover, it is often impossible to obtain large, well-distributed samples of images corresponding to certain diseases and anomalies that, while relevant, do not occur frequently \cite{Goceri2023}.
 
Two avenues for solving the lack of data in these domains are \emph{resampling} and \emph{data augmentation}. The former consists of artificially oversampling examples from low-frequency data while sometimes reducing the number of samples from high-frequency data. The latter consists of generating synthetic data for under-represented classes to even out the distribution of a dataset.

Resampling techniques have been used with relative success in several long-tailed problems, but can introduce \emph{unwanted biases} into downstream tasks \cite{8551020} and often lead to \emph{overfitting} \cite{9115249}.

Data augmentation is the natural response to these issues. It represents a booming area of research comprising several different families of algorithms, such as geometric transformations (rotations, scaling, cropping, etc.), creation of synthetic samples, mixing-based methods \cite{Chawla_2002, bao2023dpmix, yun2019cutmix} domain translation-based methods \cite{Zheng_2019} and generative methods \cite{mariani2018bagan, MORENOBAREA2020113696}.

We propose a new data augmentation method that manipulates latent space representations of images from pre-trained diffusion models, thereby generating new images to augment under-represented classes. Specific features of the data are selected via activation maps, which are then combined to produce images similar to the ones from actual data belonging to long tail classes.

The combination of latent space representations is challenging to perform through naive methods due to \emph{interference phenomena} between the post-processing of the features. We tackle this issue as a problem of \emph{compositional generalization} and apply the framework of \emph{iterated learning} (IL) \cite{ren2019compositional} with sparsified embeddings to our target data augmentation framework.

The main inspiration of IL comes from models of cultural evolution \cite{kirby2008cumulative} in which iterations of teacher-student interactions encourage useful compression and the formation of a "shared language" adapted to a task \cite{beppu2009iterated, kirby2015compression, ren2019compositional}. In particular, recently \cite{ren2023improving} have obtained favorable results related to composing distinct features when using sparsified state spaces, with a sparsification method called \emph{Simplicial Embedding} (SE)  \cite{lavoie2022simplicial}. The concept entails that if, in an IL iteration, the "teacher" version of the model is obliged to propagate a sparsified version of latent vectors to a new "student" version before training, it will impose an information bottleneck across iterations, resulting in significant improvements at downstream tasks that require compositional reasoning.

In short, we applied a new version of the IL+SE method to map the already rich latent space of a pre-trained Stable Diffusion Model \cite{rombach2022highresolution} using task-specific activation masks \cite{chu2020feature}. We propose mixing existing points in this sparse latent space to achieve better fusion at the feature level.

\section{Related Work}
\label{sec:related}
As previously mentioned, several data augmentation techniques have 
been used recently; see \cite{yang2023image} for a review. Of the methods described, the most relevant to this work are Image Mixing and Deep Generative Model approaches.
\subsection{Image mixing}
Image mixing generally consists of creating new data points by combining two or more existing ones. MixUp\cite{zhang2018mixup} and SMOTE\cite{Chawla_2002} utilize convex combinations of existing data to create new samples, with SMOTE selecting the same class neighbors as pairs. CutMix\cite{yun2019cutmix} randomly samples from the base dataset and removes patches from an image to replace with a patch from another. These methods can also be performed in the feature space.

The main problems of these methods are that they fail to produce novel, realistic samples and lack consistency in preserving labels before and after augmentation. We aim to solve the first issue by leveraging the latent space of a pre-trained stable diffusion model. For the second issue, we attempt to solve the problem using saliency methods and a separable sparse latent space.
\subsection{Deep Generative Models}
Approaches based on generative models sample new training examples through model inference. GANs \cite{goodfellow2014generative} are the most popular generative framework for data augmentation due to their fast inference and realistic generation \cite{biswas2023generative}. However, GANs are widely known to be unstable at training time, and are prone to mode collapse \cite{mode}.

Latent Diffusion Models \cite{rombach2022highresolution} have long surpassed GANs in image generation quality \cite{dhariwal2021diffusion}, and off-the-shelf models have been used to significant effect in data augmentation \cite{trabucco2023effective, Qin_2023_CVPR}. However, Diffusion Models generally suffer from slow inference speed and quality degeneration with long-tailed data \cite{Qin_2023_CVPR}. To address these issues, we work within a modified sparse version of the rich latent space defined by a trained Stable Diffusion Model, to mix existing data points instead of generating new ones from scratch. By doing this, we can create high-quality samples with very few (if any) diffusion inference steps.  
\section{Method}
\label{sec:method}
\begin{figure}
    \centering
    \includegraphics[width = \linewidth]{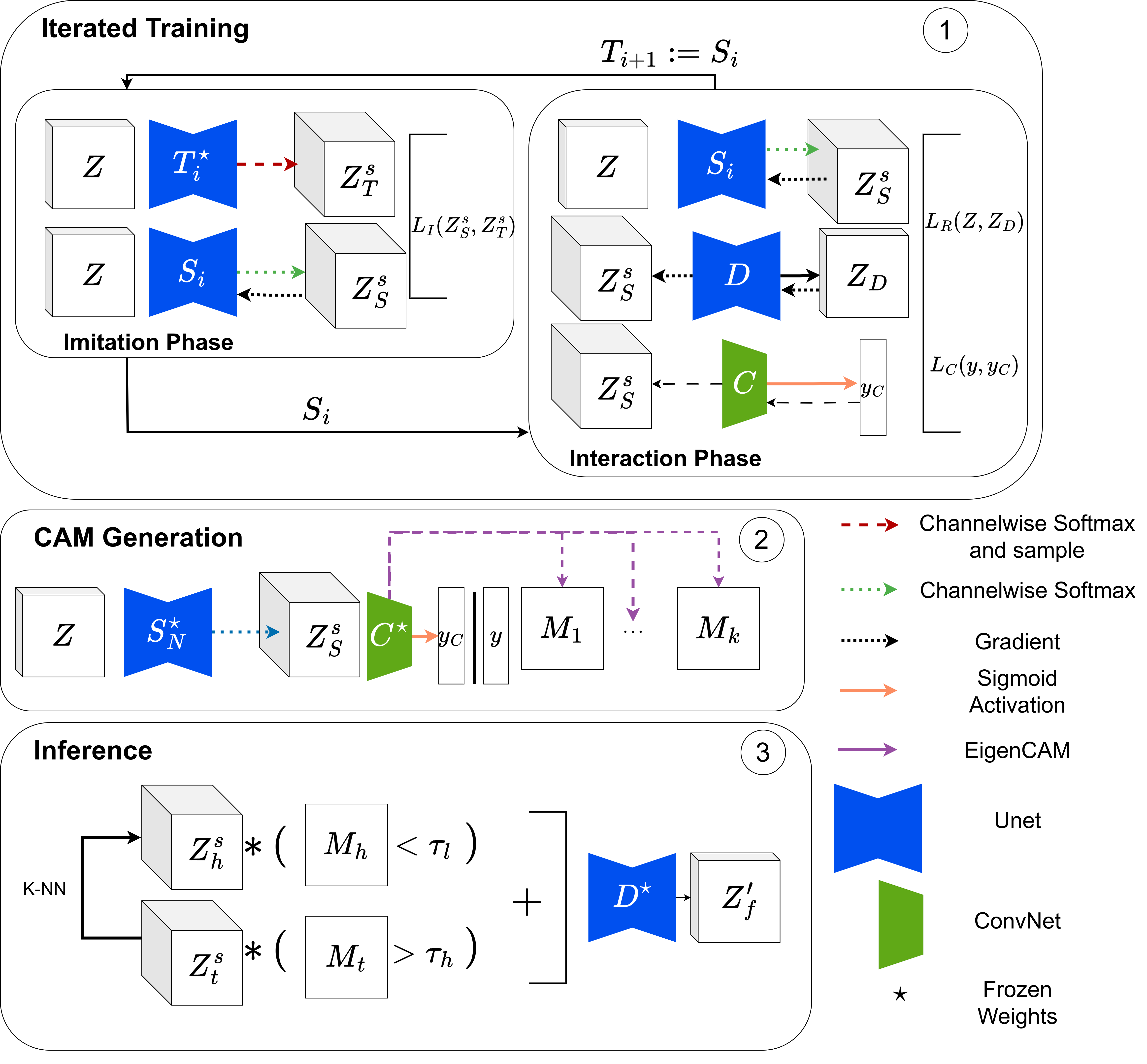}
    \caption{Proposed method: Stage 1 (Iterated training), iteratively train a student network $S_i$ to imitate a frozen teacher network $T_i$, which corresponds to the student network of the previous iteration $S_{i-1}$ in mapping the original latent vectors $Z$ to a semantically separable sparse domain $Z^s$. Also, jointly train said student with a classifier $C$ and a decoder $D$ to classify and map vectors from the sparse domain back to the original domain. In Stage 2 (CAM generation), we use EigenCAM to generate class activation maps ($M_i$ for classes I in $ [ 1,k] $) for each vector, using the classifier trained in Stage 1. Finally, in stage 3 (Inference), we find a head class near neighbor $Z_h^s$ for each tail class vector $Z_t^s$, and we combine them using their respective Class Activation Maps (CAM) as masks, taking the top activations from the tail vector and the bottom activations from the head vector. Finally, we combine these activations and pass them through $D$ to generate a new tail class vector.}
    \label{fig:Method}
\end{figure}
Our proposed method, shown in Figure \ref{fig:Method}, consists of three stages: (i) iterated training, (ii) class activation map generation, and (iii) inference. Note that each stage is applied to the latent vectors produced by a pre-trained Latent Diffusion Model, and not to the images themselves.

\begin{figure}
    \begin{subfigure}[t]{0.15\textwidth}
         \centering
         \includegraphics[width=\textwidth]{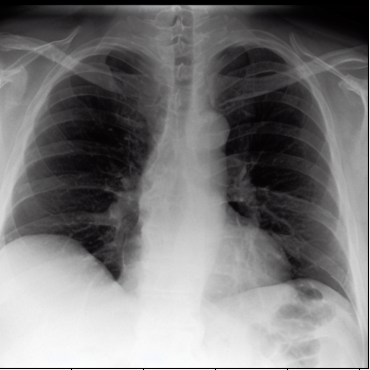}
         \subcaption{(a.1)}
         \label{tail}
    \end{subfigure}
    \hfill
    \begin{subfigure}[t]{0.15\textwidth}
         \centering
         \includegraphics[width=\textwidth]{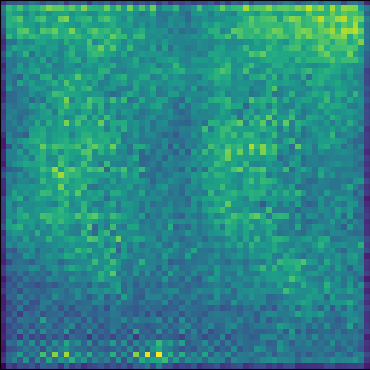}
         \caption{(a.2)}
         \label{tail}
    \end{subfigure}
    \hfill
    \begin{subfigure}[t]{0.15\textwidth}
         \centering
        \includegraphics[width=\textwidth]{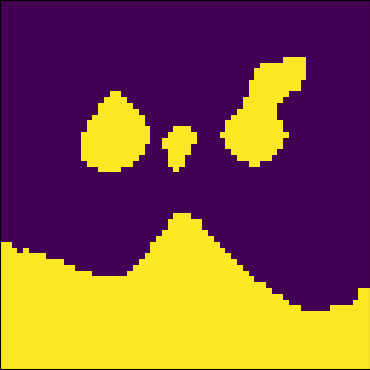}
         \caption{(a.3)}
         \label{tail}
\end{subfigure}
        \begin{subfigure}[t]{0.15\textwidth}
         \centering
         \includegraphics[width=\textwidth]{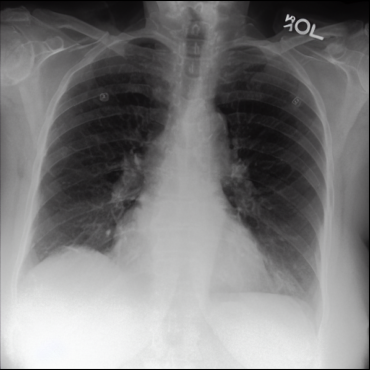}
         \caption{(b.1)}
         \label{tail}
    \end{subfigure}
    \hfill
    \hfill
    \begin{subfigure}[t]{0.15\textwidth}
         \centering
         \includegraphics[width=\textwidth]{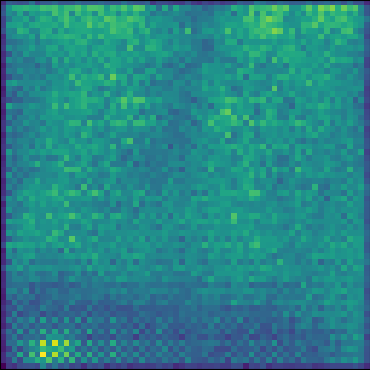}
         \caption{(b.2)}
         \label{tail}
    \end{subfigure}
    \hfill
    \begin{subfigure}[t]{0.15\textwidth}
         \centering
        \includegraphics[width=\textwidth]{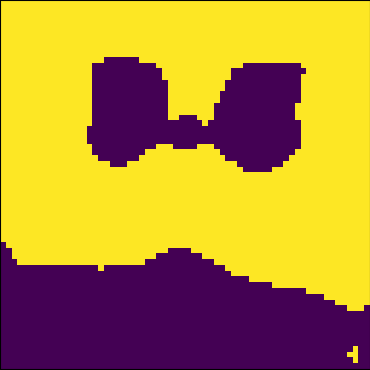}
         \caption{(b.3)}
         \label{tail}
    \end{subfigure}
    \hfill
    \begin{subfigure}[t]{0.15\textwidth}
         \centering
         \includegraphics[width=\textwidth]{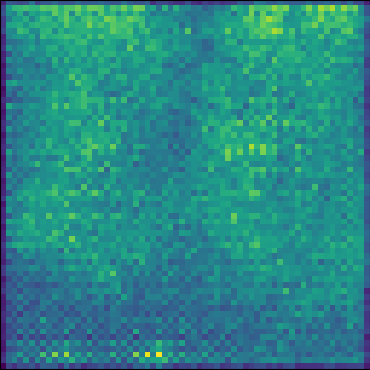}
         \caption{(c)}
         \label{tail}
    \end{subfigure}
    \hfill
    \hfill
    \begin{subfigure}[t]{0.15\textwidth}
         \centering
         \includegraphics[width=\textwidth]{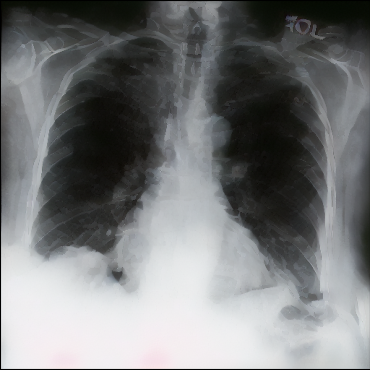}
         \caption{(d)}
         \label{tail}
    \end{subfigure}
    \hfill
    \begin{subfigure}[t]{0.15\textwidth}
         \centering
         \includegraphics[width=\textwidth]{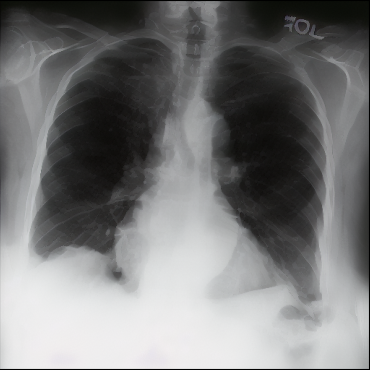}
         \caption{(e)}
         \label{tail}
    \end{subfigure}
    \hfill
\caption{Fusion process applied to an image from the tail class Tortuous Aorta (a.1) and one of its neighbor images from the head class Atelectasis (b.1). (a.2) and (b.2) are channelwise Maximum Intensity Projections of the sparse vectors obtained from (a.1) and (b.1) respectively. In (a.3) and (b.3), we use EigenCAM to find attention maps for each sparse vector and define binary masks (yellow is one and dark purple is zero) using $\tau_h = \tau_l = 0,4$ as thresholds. We combine the masked sparse vectors into (c) and decode the vector into a fused image (d). Finally, we apply five inference steps in (e) to obtain a less noisy image.}
 \label{fig:Example}
\end{figure}
\subsection{Iterated training}
In this stage, we learn a translation from the diffusion latent space to a sparse high dimensional representation \cite{lavoie2022simplicial}, while training a convolutional classifier for this space.  

To do this, we implement a student-teacher \cite{abbasi2019modeling} training regime, which iteratively runs through two phases: imitation and interaction.

We train a student network $S_i:\theta \times Z\rightarrow Z^s_S$ during the imitation phase from scratch. This network transforms latent vectors $Z \subseteq \mathcal{R}^{H \times W \times C }$ into sparse high dimensional vectors $Z^s_S \subseteq  \mathcal{R}^{H \times W \times C' }$, with $C < C'$. We enforce sparsity by applying a channel-wise softmax to these output vectors, as follows:
\begin{equation}
\begin{aligned}
z^S_{ijk} := \frac{exp(S_i(z_{ijk}))}{\sum^{C'}_{\hat{k} = 1} exp(S_i(z_{ij\hat{k}}))}.
\end{aligned}
\end{equation}
We train $S_i$ to imitate a teacher network $T_i$, which is the student of the previous iteration, $T_i = S_{i-1}$. Instead of imitating $T_i$ directly, we use the channel-wise probabilities coded in $T_i(z^S)$ to sample a binary vector $z^S_b$, where each spatial coordinate is a one-hot vector of dimension $C'$. Minimizing the imitation loss then becomes a multilabel classification problem: 

\begin{equation}
\begin{aligned}
L_I = BCE(z^S_b, S_i(z^S))
\end{aligned}
\end{equation}
During the interaction phase, we minimize losses related to two other networks. First, we aim to make the sparse vectors good representations of the original latent vectors. To this end, we jointly train a network $D:\theta \times Z'\rightarrow Z$ with $S_i$, to minimize the reconstruction loss: 
\begin{equation}
\begin{aligned}
L_R = MSE(z, D(S_i(z_s)).
\end{aligned}
\end{equation}
Our objective entails ensuring that the sparse vectors are easily separable in their respective image classes; with this in mind, we trained a classifier $C:\theta \times X'\rightarrow Y$ to minimize the multilabel classification loss: 
\begin{equation}
\begin{aligned}
L_C = (y, C(S_i(z))).
\end{aligned}
\end{equation}
We couple the two losses via a new hyperparameter $\lambda$, and the interaction phase translates into the optimization:

\begin{equation}
\begin{aligned}
\min_{\theta_{S_i},\theta_D, \theta_C}  \EX_{z, y\in Z}[\lambda L_R +  (1-\lambda) L_C].
\end{aligned}
\end{equation}
\subsection{Class Activation Map Generation}
In this stage, we use the classifier $C$ from the previous stage to generate simple and interpretable activation maps for each class $c\in K$, to select relevant or non-relevant coordinates for classification as $c$.

First, based on the classifier $C$, we define class activation maps $M_c \subseteq\{0, 1\}^{H \times W}$ for every sparse vector $z^s\in Z^s$ and every class $c \in K$, where $M_c \approx 1$ defines which spatial coordinates of the sparse vectors are important for classification as class $c$. The CAMs are generated with EigenCAM \cite{Muhammad_2020}, which is adapted to the task because it does not require correct classification to generate attention maps. 

Next, following \cite{chu2020feature}, we separate class-specific and class-generic features by a threshold method, as follows. With upper threshold value $\tau_h < 1$ and lower value $\tau_l > 0$, set
\begin{equation}
\begin{aligned}
M^S_{c} := sgn(M_c - \tau_h)\quad 
M^G_{c} := 1 - sgn(\tau_l - M_c),
\end{aligned}
\end{equation}
whose interpretation is that $M^S_{c}$ contains coordinates of $z^s$ relevant to classify the vector as $c$, while $M^G_{c}$ contains  non-relevant coordinates.
\subsection{Inference}
In this stage, we generate new samples from tail classes. Let $H$ and $T$ be sets of head and tail classes. We first find the head class $H_c$ of highest confusion for a given tail class, provided by the highest estimated value index in the $C$ classifier prediction for the head classes. We then sample an element for each tail class example from their highest confusion head class. 

A possible naive approach would have been to create a fused vector by adding the class-specific tail features with the class generic head features, using their respective masks. However, note that the class-specific mask for the first vector and the class-generic mask for the second vector might overlap or leave empty spaces in the final vector. For these uncovered and overlapping parts of the vectors, it is not clear how to update their values in an interpretable manner. To solve this, whenever both masks $M_c^S, M_c^G$ are 0 or both are 1, we choose which vector to use at random, introducing a random mask $M^R$. The values of the maps also need to broadcasted across $C'$ channels as follows: for $M\in\{0,1\}^{H\times W}$, let $\overline{M}\in\{0, 1\}^{H \times W \times C'}$ be the tensor given by repeating $C'$ times the values of $M$. Then, define
\begin{eqnarray}
M^R &\sim& \left[\mathsf{Unif}(\{0, 1\})\right]^{H\times W},\\
z^s_R &:=& z^s_t\odot\overline{M^R} + z^s_h(\mathbf{1}-\overline{M^R}), \\
\widetilde z^s_R &:=&  z^s_R\odot(\mathbf{1}-\overline{M^G_{c}} - \overline{M^S_{c}} + 2\overline{M^G_{c}} \odot \overline{M^S_{c}}),\\
\widetilde z^{s}_t&:=&z^s_t \odot
(\overline{M^S_{c}} \odot (\mathbf{1}-\overline{M^G_{c}})),\\
\widetilde z^{s}_h&:=&z^s_h \odot
(\overline{M^G_{c}} \odot (\mathbf{1}-\overline{M^S_{c}})),\\
z^s_F&:=&\widetilde z^{s}_h + \widetilde z^{s}_t + \widetilde z^s_R.
\end{eqnarray}
Having defined the sparse fusion vector $z^s_F$, we translate it back into the base latent space $Z$ using the decoder network $D$ trained in stage 1:
\begin{equation}
\begin{aligned}
z_F:=D(z^s_F)
\end{aligned}
\end{equation}
Our results suggest that, for fast data augmentation, this approach tends to suffice; however, for high-fidelity image generation, an extra step is added, as this vector does not necessarily lie within the pre-trained VAE domain. Using the frozen pre-trained Unet $U^*$ from the Stable Diffusion Model, conditioned on the class names associated with the tail class image, the vector can be translated into the VAE domain. We take $N/d$ denoising steps, where $N$ is the number of inference steps in the original Stable Diffusion model, and $d\in[1, N]$.

\section{Results}
\label{sec:results}
The method was tested by generating tail classes of a sampled version of MIMIC-CXR-LT 2023 \cite{Holste_Wang_Jaiswal_Yang_Lin_Peng_Wang_2023, Goldberger_Amaral_Glass_Hausdorff_Ivanov_Mark_Mietus_Moody_Peng_Stanley_2000}, a multilabel multiclass long-tailed dataset of chest X-rays. Our reduced sample version has only five head classes and five tail classes. 

We compare our method with SMOTE \cite{Chawla_2002} and RoentGEN \cite{chambon2022roentgen}, which is used as our Stable Diffusion Model, on image generation quality using FID \cite{heusel2018gans}. We also evaluate using no extra inference steps, one step, and five. We train a Densenet121 \cite{huang2018densely} model using each augmented dataset to classify our sample train set and test it on a separate test set from the same challenge. We evaluate the Mean Average Precision for head and tail classes for each described method and compare them with a classifier trained on our unaugmented sample set.
\begin{table}[h]
  \centering
  \begin{tabular}{l@{}c@{}c@{}r@{}}
    \toprule
    Model & Avg Tail FID$\downarrow$ & Head mAP$\uparrow$ & Tail mAP$\uparrow$ \\
    \midrule
    Baseline & - & 0.618 & 0.155 \\
    \midrule
    SMOTE & 171.864 & 0.578 & 0.151\\
    RoentGEN@75 & 138.963 & \textbf{0.618} & \textbf{0.152}\\
    Ours@0 & 191.873 & 0.607 &  \textbf{0.152}\\
    Ours@1 & 191.646 & 0.595 & 0.143 \\
    Ours@5 & \textbf{130.110} & 0.595 & 0.144  \\
    \bottomrule
  \end{tabular}
  \caption{Results measuring FID \cite{heusel2018gans} and Mean Average Precision for tail and head classes. We use 75 inference steps for RoentGEN, following \cite{chambon2022roentgen}, and 0, 1, and 5 steps for our method.} 
    \label{tab:FID}
\end{table}
  As \cref{tab:FID} shows, we obtain lower image quality than the other methods when using 0 and 1 inference steps. However, we get a lower FID with only five inference steps than RoentGEN using 75 inference steps. 
  Unexpectedly, all tested augmentation methods have worse results when used to augment a classifier than the baseline. 
  One possible explanation for this behavior is how labels are assigned to new data: As a multi-labeled classification problem, our approach assigns labels from the corresponding head and tail images to the fused latent vectors. However, this may introduce biases that lead to misclassification of real samples. Furthermore, despite the good FID performance, our method performs noticeably worse in the classification task (mAP). This might be a consequence of the diffusion process, which does not guarantee maintaining the existing labels in the fused latent vector.

\section{Conclusion}
\label{sec:conclusion}
In this work, we present a novel method for data augmentation and data generation on long-tailed datasets. By leveraging pre-trained Latent Diffusion Models, compositional learning, and saliency methods, we generate new examples of underrepresented classes. We give a detailed mathematical description of our method and run experiments on image generation and data augmentation in the medical domain for multi-label classification using a small subset of MIMIC-CXR-LT \cite{Holste_Wang_Jaiswal_Yang_Lin_Peng_Wang_2023, Goldberger_Amaral_Glass_Hausdorff_Ivanov_Mark_Mietus_Moody_Peng_Stanley_2000}. 

Using the Latent Diffusion Model RoentGEN \cite{chambon2022roentgen} to run five inference steps on our generated vectors, we obtain competitive results in image generation quality. Counter-intuitively, we found that using more inference steps negatively impact the downstream image classification when using our augmented data. This might be a consequence of our approach not necessarily maintaining data labels throughout the diffusion process.



In future steps, we plan to experiment with larger datasets and different techniques for assigning labels and compare our results with a broader range of generation and augmentation methods across a more comprehensive range of tasks. 

\section{Acknowledgements}
\label{sec:acknowledgements}

This work was supported by ANID Chile, Fondecyt Regular grant 1231724, as well research centers of excellence with code FB210017 (Basal CENIA), ICN2021\_004 (Millenium iHealth), and ICN17\_002 (Millenium IMFD).

{\small
\bibliographystyle{ieeenat_fullname}
\bibliography{11_references}
}

\ifarxiv \clearpage \appendix \section{Appendix Section}
\label{sec:appendix_section}
Supplementary material goes here.
 \fi

\end{document}


\title{\paperTitle}
\author{\authorBlock}
\maketitlesupplementary

\appendix
\section{Appendix Section}
\label{sec:appendix_section}
Supplementary material goes here.

{\small
\bibliographystyle{ieeenat_fullname}
\bibliography{11_references}
}